\documentclass[twoside,11pt]{article}

\usepackage{jmlr2e}
\usepackage{bbm}
\usepackage{amsfonts}
\usepackage{amsmath}
\usepackage{bm}
\usepackage{hyperref}
\hypersetup{hypertex=true,
            colorlinks=true,
            linkcolor=blue,
            anchorcolor=blue,
            citecolor=blue}

\jmlrheading{}{}{}{}{}{Xiao Fang and Malay Ghosh}

\ShortHeadings{Posterior Consistency for Bayesian Relevance Vector Machines}{Fang  and Ghosh}
\firstpageno{1}

\begin{document}

\title{Posterior Consistency for Bayesian Relevance Vector Machines}

\author{\name Xiao Fang \email xiaofang@ufl.edu\\
       \name Malay Ghosh \email ghoshm@ufl.edu \\
       \addr Department of Statistics\\
       University of Florida\\
       Gainesville, FL 32611, USA}

\editor{}

\maketitle

\begin{abstract}
Statistical modeling and inference problems with sample sizes substantially smaller than
the number of available covariates are challenging. \citet{Chakraborty:2012} did a full hierarchical Bayesian analysis of nonlinear regression in such situations using relevance vector machines based on reproducing kernel Hilbert space (RKHS). But they did not provide any theoretical properties 
associated with their procedure. The present paper revisits their problem,  introduces a 
new class of global-local priors different from theirs, and provides results on posterior consistency as well
as posterior contraction rates. 
\end{abstract}

\section{ Introduction}
\noindent
Regression techniques are widely used virtually in any field 
demanding quantitative analysis. Even until today, much of this 
analysis relies on a linear relationship between the predictors
and the response variables. This, however, is often more a
convenience than reality. There is no dearth of problems of
applied interest where the linearity assumption fails, and
non-linear regression is called for. Fortunately, recent
advancement in computer capability has allowed statisticians to
tackle such non-linear regression problems. In addition, 
statisticians are now able to handle data where the number of
covariates (say, $p$) far exceeds the sample size (say, $n$), 
a situation of natural 
ocurrence, for example in microarray experiments, image analysis,
and a variety of commonly encountered problems in medicine, business,
economics, sociology and others.\\

\noindent
\citet{Chakraborty:2012} considered one such problem
arising from near infrared (NIR) spctroscoy where spectral measurements
typically produce many more covariates (wavelets, channels) than 
calibration measurements (samples). They considered a full hierarchical
Bayesian analysis of such data using relevance vector machines (RVM's).
RVM's are machine learning techniques, originally introduced by \citet{Tipping:2000, Tipping:2001}
 and \citet{Bishop:2000}. These authors essentially
used an empirical Bayes procedure involving Type II maximum likelihood
\citep{Good:1965} estimators of prior parameters. Unlike them,\citet{Chakraborty:2012} used a hierarchical Bayesian procedure by assigning distributions
to the prior parameters. Hierarchical Bayes procedures typically hold
advantage over empirical Bayes procedures in that unlike the latter, 
they can model the uncertainty in estimating the prior parameters,
thus particularly useful for prediction.\\

\noindent
The RVM regression approach of \citet{Chakraborty:2012}. was based on
reproducing kernel Hilbert space (RKHS). 
While they could implement their procedure via
Markov chain Monte Carlo (MCMC),  
they did not establish any
theoretical properties of their method. The basic objective
of this paper is to provide theoretical underpinnings to the problem introduced by
 \citet{Chakraborty:2012}.  We have introduced instead a class of global-local priors different from the one of \citet{Chakraborty:2012}. Global-local priors are widely used in high-dimensional statistics, for example, by \citet{Carvalho:2010} , \citet{Polson:2010} and many others. One of the attractive features of our
 priors is that they can handle both sparse and dense situations, and the asymptotics is based on
 the sample size $ n$ tending to infinity. \\
\\
Our paper essentially consist of two parts. In the first part of this paper, we have proved under minimal assumptions posterior consistency as well as posterior  contraction rate for a bounded kernel which includes the well-used Gaussian kernel under some mild conditions.  As mentioned, the results are very general where the number of covariates can far exceed the sample size n. The prior used is a certain class of global-local  priors, and the global parameter plays a key role in establishing posterior consistency as well as posterior contraction. With appropriate choice of  this parameter, we are able to obtain asymptotic minimax posterior  contraction rate as well. The second part of the paper deals with polynomial kernels where we are able to establish posterior consistency as well as posterior
contraction rates.\\

\noindent
The outline of the remaining sections is as follows. We have introduced the
hierarchical Bayesian model in Section 2 for bounded kernels and have derived the marginal
posterior of the regression parameter of interest.
Section 3 deals with the bounded kernel  and  posterior consistency and contraction are established under the proposed  model. Section 4 deals with results involving polynomial kernels with
fixed kernel parameter. Some final discussions are made in Section 5. \\

\section{Hierarchical Regression Model Based on RKHS}
In this section we introduce the reproducing kernel Hilbert space (RKHS) and hierarchical Bayesian model based on RKHS.
\subsection{Regression Model Base on RKHS}
For a regression model, we have a training set $\{\bm{Y}_{in},\bm{x}_i\}, i=1,2,\cdots,n$, where
$\bm{Y}_{in}$ is the response variable and $\bm{x}_i=(x_{i1},\cdots,x_{ip})^T$ is the vector of covariates of size p corresponding to $\bm{Y}_{in}$. Given the training data our goal is to find an appropriate function $f(\bm{x})$ to predict the response $y$ in the test set based on the covariates $\bm{x}$. This can be viewed as a regularization problem of the form
\begin{equation}
min_{f \in \mathbb{H}}[\Sigma_{i=1}^{n}L(\bm{Y}_{in},f(\bm{x}_i))+\lambda J(f)] \label{minRKHS1}
\end{equation}
where $L(y,f(x))$ is a loss function, $J(f)$ is a penalty functional, $\lambda > 0$ is the smoothing parameter, and $\mathbb{H}$ is a space of functions on which J(f) is defined. In this article, we consider $\mathbb{H}$ to be a reproducing kernel Hilbert space (RKHS) with
kernel $K$, and we denote it by $\mathbb{H}_K$ . A formal definition of RKHS in given in  \citet{Aronszajn:1950}, \citet{Parzen:1970} and \citet{Wahba:1990}.\\
\\
For an $h \in \mathbb{H}_K$, if $f(\bm{x})=h(\bm{x})$, we take $J(f)=\parallel h \parallel_{\mathbb{H}_K}$ and rewrite (\ref{minRKHS1}) as 
\begin{equation}
min_{h \in \mathbb{H}_K}[\Sigma_{i=1}^{n}L(\bm{Y}_{in},h(\bm{x}_i))+\lambda \parallel h \parallel_{\mathbb{H}_K}]. \label{minRKHS2}
\end{equation}
The estimate of $f$ is obtained as a solution of (\ref{minRKHS2}). It can be shown that the solution is finite-dimensional and leads to a representation of $f$  as 
\begin{equation}
f(\bm{x}_i)=\Sigma_{j=1}^{n}\beta_jK(\bm{x}_i,\bm{x}_j|\theta). \label{RKHS formula}
\end{equation} 
\\
Two classical choices of the reproducing kernel $K$ are\\
(a) The Gaussian kernel $K(\bm{x}_i,\bm{x}_j)=exp\{-\lVert\bm{x}_i-\bm{x}_j \rVert^2/\theta \}$, $\theta >0$,    \\
(b) The polynomial kernel $K(\bm{x}_i,\bm{x}_j)= (\bm{x}_i \cdot\bm{x}_j+1)^\theta$, $\theta > 0$.\\
\\

\subsection{Hierarchical Bayes Relevance Vector Machine}

 Assume the true model is $\bm{Y}_n = \bm{K}_{n} \bm{\beta}_{0n}+ \bm{\epsilon}_n$. Here $\bm{Y}_{n}=(\bm{Y}_{1n},\cdots,\bm{Y}_{nn})^T$, $\bm{\epsilon}_n \sim N(0,\sigma_0^2\bm{I}_n)$, $(\bm{K}_{n})_{ij}=K(\bm{x}_i,\bm{x}_j|\theta)$, $i,j=1,2,\cdots,n$, $\bm{\beta}_{0n}=(\beta_{01},\cdots,\beta_{0n})^T$, $\bm{\epsilon}_n \sim N(0,\sigma_0^2\bm{I}_n)$. Let $\bm{X}_n^T=(\bm{x}_1,...,\bm{x}_n)$.\\
\\  
  We consider the hierarchical model as follows: $\bm{Y}_{in}|\theta,\bm{X}_n,\bm{\beta}_n,\sigma^2 \stackrel{ind}\sim N(\bm{K}_{in}^T\bm{\beta}_n,\sigma^2) $ 
with 
$\bm{K}_{in}^T=(K(\bm{x}_i,\bm{x}_1|\theta),\cdots,K(\bm{x}_i,\bm{x}_n|\theta))$, $i=1,\cdots,n$. \\
\\
The following  hierarchical is assigned to unknown parameters $\bm{\beta}_n$, $\sigma^2$:\\
\textbf{Model 1:}\\
(i)  $ \bm{\beta}_n|\sigma^2,\bm{\Lambda_n}^2 \sim N(\mathbf{0},\sigma^2 \tau_n^2 \bm{\Lambda}^2_n),\bm{\Lambda}_n^2=diag(\lambda_1^2,...,\lambda_n^2)$ ,   \\
(ii)  $\sigma^2 \sim IG(a/2,b/2)$ , \\
(iii) $\lambda_i^2 \stackrel{i.i.d}{\sim} \pi(\lambda_i^2)$ .\\
\\
\begin{remark}
 Here we assign global local shrinkage prior to  the coefficient $\bm{\beta}_n$, and the parameter $\tau_n^2$ is called the global shrinkage parameter.   Global local shrinkage prior is  widely used   in high dimensional regression problem nowadays, and it can leads to posterior consistency, see \citet{Ghosh:2017}, \citet{Pas:2014} and \citet{Song:2017}. Our model is essentially a linear model, and $\bm{K}_{n}$ in our case becomes the  design  matrix with coefficient $\bm{\beta}_n$.  We need to  add some regularization conditions on $\bm{K}_{n}$  and also on the prior distributions of $\lambda^2_i$. Our model is similar to that of  \citet{Ghosh:2017}, but we assume $\sigma^2$ is unknown and $\bm{K}_{n} \neq \bm{I}_{n}$, which makes our analysis more complicated.
\end{remark}
\noindent
With these priors we  get 
\begin{equation}
\begin{split}
&\pi(\bm{\beta}_n,\sigma^2,\bm{\Lambda}^2_n|\bm{Y}_n,\bm{X}_n) \\
\propto & (\sigma^2)^{-n-a/2-1} \pi(\bm{\Lambda}^2_n){|\bm{\Lambda}_n^2|}^{-1/2}
\exp [\frac{-b-{\bm{\beta}_n}^T{(\tau_n^{-2}\bm{\Lambda}^{-2}_n)}{\bm{\beta}_n}-(\bm{Y}_n-\bm{K}_n
\bm{\beta}_n)^T(\bm{Y}_n-\bm{K}_n\bm{\beta}_n)}{2\sigma^2}];
\end{split} \label{posterior 1}
\end{equation}
 
\begin{equation}
\bm{\beta}_n|\sigma^2,\bm{\Lambda}^2_n,\bm{Y}_n,\bm{X}_n 
\sim N((\bm{K}_n^2+\tau_n^{-2}\bm{\Lambda}^{-2}_n)^{-1}\bm{K}_n\bm{Y}_n,\sigma^2(\bm{K}_n^2+\tau_n^{-2}\bm{\Lambda}^{-2}_n)^{-1}); \label{posterior 2}
\end{equation}

\begin{equation}
\begin{split}
&\pi(\sigma^2,\bm{\Lambda}^2_n|\bm{Y}_n,\bm{X}_n)  \\
\propto &{(\sigma^2)^{-n/2-a/2-1}} \pi(\bm{\Lambda}^2_n){|\bm{\Lambda}^2_n|}^{-1/2}
{|\bm{K}_n^2+\tau_n^{-2}\bm{\Lambda}^{-2}_n|}^{-1/2}
 \exp[\frac{-b-{\bm{Y}_n}^T{(\bm{I}_n-\bm{K}_n(\bm{K}_n^2+\tau_n^{-2}\bm{\Lambda}^{-2}_n)^{-1}\bm{K}_n)}{\bm{Y}_n}}{2\sigma^2}];
 \end{split}\label{posterior 3}
\end{equation}

\begin{equation}
\begin{split}
&\pi(\bm{\Lambda}^2_n|\bm{Y}_n,\bm{X}_n) \\
\propto &  \pi(\bm{\Lambda}^2_n){|\bm{K}_n^2\bm{\Lambda}_n^{2}+\tau_n^{-2}\bm{I}_n|}^{-1/2}
{(b+{\bm{Y}_n}^T{[\bm{I}_n-\bm{K}_n(\bm{K}_n^2+\tau_n^{-2}\bm{\Lambda}^{-2}_n)^{-1}\bm{K}_n]}{\bm{Y}_n})}^{-n/2-a/2};
\end{split}\label{posterior 4}
\end{equation}

\begin{equation}
\begin{split}
&E(\bm{K}_n\bm{\beta}_n|\bm{Y}_n,\bm{X}_n)-\bm{K}_n\bm{\beta}_{0n}\\
=&E(E(\bm{K}_n\bm{\beta}_n|\sigma^2,\bm{\Lambda}^2_n,\bm{Y}_n,\bm{X}_n)|\bm{Y}_n,\bm{X}_n)-\bm{K}_n\bm{\beta}_{0n}\\
=& E[\bm{K}_n(\bm{K}_n^2+\tau_n^{-2}\bm{\Lambda}^{-2}_n)^{-1}\bm{K}_n(\bm{Y}_n-\bm{K}_n\bm{\beta}_{0n})|\bm{Y}_n,\bm{X}_n] \\
& -E[\bm{K}_n(\bm{K}_n^2+\tau_n^{-2}\bm{\Lambda}^{-2}_n)^{-1}\tau_n^{-2}\bm{\Lambda}^{-2}_n\bm{\beta}_{0n}|\bm{Y}_n,\bm{X}_n].
\label{difference of beta}
\end{split} 
\end{equation}

\subsection{Notations}
For a vector $\bm{v}\in \mathbf{R}^n$, $\parallel v\parallel_{2,n}=(\Sigma_{i=1}^{n} {v_i}^2)^{\frac{1}{2}}$ denote the $l_2$ norm. Let $\lambda_{max}= max \{\lambda_1,\cdots,\lambda_n \}$, $\lambda_{min}= min\{\lambda_1,\cdots,\lambda_n \}$, where $\lambda_1,\cdots,\lambda_n$ are the diagonal elements of $\bm{\Lambda}_n$.
$E_0$ denotes expectation under true model, i.e $E_{\bm{\beta}_{0n}}$.

\section{Hierarchical Bayesian Model with Bounded  Kernel}
In this section, we consider the case where model 1   has bounded  kernel with fixed  parameter $\theta$. Before studying the property of posterior distribution, we state some regularity conditions on the matrix $K_n$ and the  true parameters  $\bm{\beta}_{0n}$, $\sigma^2_0 $ .\\
\\
\noindent
\textbf{Regularity conditions:}\\
 \textbf{(A1)} the design matrix $\bm{X}_n$ satisfies  
\begin{displaymath}
c_1 \bm{I}_n \leq \bm{K}_n\leq c_2 \bm{I}_n
\end{displaymath}
for sufficiently large $n$, where $c_1, c_2 >0$ do not depend on n.\\
\textbf{(A2)}  $|\beta_{0n,i}|<M$,  $M $ and $\sigma^2_0 $  does not depend on n.\\
\textbf{(A3)}  Let $q_n$ be the number of nonzero elements in $\bm{\beta}_{0n}$, $q_n=o(n)$.\\
\\
\begin{remark} 
(A1) states that the kernel $\bm{K}_n$ is bounded. For Gaussian kernel with fixed parameter $\theta >0$, if  the design matrix $\bm{X}_n$ satisfies the orthogonality condition, namely,
 $\bm{X}_n\bm{X}_n^T=p\bm{I}_n$, $p>n$, then $K_n-I_n\rightarrow 0$ as $n\rightarrow\infty$ so that  $\frac{1}{2} \bm{I}_n\le \bm{K}_n \le 2 \bm{I}_n$ for sufficiently large $n$. Actually we can extend  conditions on $\bm{X}_n $ such that the Gaussian kernel $\bm{K}_n$ still satisfies (A1):
\begin{lemma}
If $K_n$ is a Gaussian  kernel with parameter $\theta$ and $\lVert\bm{x}_i-\bm{x}_j \rVert^2 \ge k(n) = 2\theta\log n $ for sufficiently large $n$ if $i \neq j$,
 then there exists $N >0$ such that when $n>N$, $(1-\frac{1}{n})\bm{I}_n \le \bm{K}_n \le (1+\frac{1}{n})\bm{I}_n$.
\end{lemma}
\textbf{Proof:} It suffices to show that  for every $\bm{c} \neq \bm{0}$, $\bm{c}^T\bm{K}_n\bm{c} \le (1+\frac{1}{n})\bm{c}^T\bm{c}$ for large $n$.
But 
\begin{displaymath}
\begin{split}
&\bm{c}^T\bm{K}_n\bm{c} \\
\le & \Sigma_{i=1}^{n}{c_i}^2+\Sigma_{1\le i \neq j \le n}|c_i||c_j|/\exp(k(n)/\theta) \\
= & (\Sigma_{i=1}^{n}{c_i}^2+[(\Sigma_{i=1}^{2}{|c_i|})^2-\Sigma_{i=1}^{2}{c_i}^2]/(2\exp(k(n)/\theta))\\
\le & [1-\frac{1}{2\exp(k(n)/\theta)}]\Sigma_{i=1}^{n}{c_i}^2
+\frac{n\Sigma_{i=1}^{2}{c_i}^2}{2\exp(k(n)/\theta)}\\
\le &[1+\frac{n-1}{2\exp(k(n)/\theta)}]
\Sigma_{i=1}^{n}{c_i}^2\\
\le &  (1+\frac{1}{n})\bm{c}^T\bm{c},
\end{split}
\end{displaymath}
for sufficiently large $n$. Similarly, we have $\bm{c}^T(p^{-\theta}\bm{K}_n)\bm{c} \ge (1-\frac{1}{n})\bm{c}^T\bm{c}$.\\
\end{remark}
\noindent
\begin{remark}  
Compared to \citet{Ghosh:2017},  who assumed $K_n=I_n$, condition(A1)  requires only boundedness of  $\bm{K}_n$ in both directions.\citet{Ghosh:2017} also do not impose any assumption  to the true parameter $\bm{\beta}_{0n}$, while  in condition (A2), we assume it is bounded. In consequence, our posterior contraction rate can be faster than the minimax rate $q_n \log(\frac{n}{q_n})$,  as we will now demostrate in the following theorems.\\ 
\end{remark}
\noindent
\\
\begin{theorem} 
Let $K_n$ be a kernel with fixed parameter, and conditions (A1)-(A3) hold.  Consider the priors assigned to $\bm{\Lambda}_n$ and $\sigma^2$ in Section 2.2. Then if $\tau^2_n \preceq n^{-\frac{3}{2}}q_n$ and
\begin{displaymath}
\int \lambda^{4} \pi(\lambda^2)  d{\lambda^2} < \infty,
\end{displaymath}
  
\begin{displaymath}
 E_0\parallel E(\bm{K}_n \bm{\beta}_{n}|\bm{Y}_n,\bm{X}_n)-\bm{K}_n \bm{\beta}_{0n}\parallel_{2,n}^2 \preceq q_n
\end{displaymath}
as $n \to \infty.$ \label{theorem1}
\end{theorem}
\begin{remark} 
This theorem holds for both $p \leq n$ and $p>n$ case.
\end{remark}
\noindent
\\
\textbf{Proof of Theorem \ref{theorem1}:}  Denoting the rightmost side of (\ref{difference of beta}) as $I-II$, it suffices to show that $E_0 \parallel I \parallel_{2,n}^2 \preceq q_n$  and $E_0\parallel II \parallel_{2,n}^2 \preceq q_n$  as $n \to \infty$.\\
\\
First, we prove $E_0 \parallel I \parallel_{2,n}^2 \preceq q_n$. It suffices to show that
\begin{displaymath}
E_0 E[\|\bm{K}_n(\bm{K}_n^2+\tau_n^{-2}\bm{\Lambda}^{-2}_n)^{-1}\bm{K}_n(\bm{Y}_n-\bm{K}_n\bm{\beta}_{0n})\|_{2,n}^2|\bm{Y}_n,\bm{X}_n] \preceq q_n .
\end{displaymath}
\noindent
To this end, we proceed as follows :
\begin{scriptsize}
\begin{equation}
\begin{split}
& E[\|\bm{K}_n(\bm{K}_n^2+\tau_n^{-2}\bm{\Lambda}^{-2}_n)^{-1}\bm{K}_n(\bm{Y}_n-\bm{K}_n\bm{\beta}_{0n})\|_{2,n}^2|\bm{Y}_n,\bm{X}_n]\\
=& E[(\bm{Y}_n-\bm{K}_n\bm{\beta}_{0n})^T \bm{K}_n(\bm{K}_n^2+\tau_n^{-2}\bm{\Lambda}^{-2}_n)^{-1}\bm{K}_n^2   (\bm{K}_n^2+\tau_n^{-2}\bm{\Lambda}_n^{-2})^{-1}  \bm{K}_n(\bm{Y}_n-\bm{K}_n\bm{\beta}_{0n})|\bm{Y}_n,\bm{X}_n]\\
\leq & c_2^2  E[(\bm{Y}_n-\bm{K}_n\bm{\beta}_{0n})^T \bm{K}_n(\bm{K}_n^2+\tau_n^{-2}\bm{\Lambda}^{-2}_n)^{-2} \bm{K}_n(\bm{Y}_n-\bm{K}_n\bm{\beta}_{0n})|\bm{Y}_n,\bm{X}_n]  \\
= & c_2^2  E[(\bm{Y}_n-\bm{K}_n\bm{\beta}_{0n})^T \bm{K}_n(\bm{K}_n^2+\tau_n^{-2}\bm{\Lambda}^{-2}_n)^{-1/2}(\bm{K}_n^2+\tau_n^{-2}\bm{\Lambda}^{-2}_n)^{-1}(\bm{K}_n^2+\tau_n^{-2}\bm{\Lambda}^{-2}_n)^{-1/2} \bm{K}_n(\bm{Y}_n-\bm{K}_n\bm{\beta}_{0n})|\bm{Y}_n,\bm{X}_n]  \\
\leq & (c_2^2 /c_1^2) E[(\bm{Y}_n-\bm{K}_n\bm{\beta}_{0n})^T \bm{K}_n(\bm{K}_n^2+\tau_n^{-2}\bm{\Lambda}^{-2}_n)^{-1} \bm{K}_n(\bm{Y}_n-\bm{K}_n\bm{\beta}_{0n})|\bm{Y}_n,\bm{X}_n]  \\
\leq &  \tau_n^2(c_2^2 /c_1^2) E[\lambda^{2}_{max}(\bm{Y}_n-\bm{K}_n\bm{\beta}_{0n})^T \bm{K}_n \bm{K}_n(\bm{Y}_n-\bm{K}_n\bm{\beta}_{0n})|\bm{Y}_n,\bm{X}_n]  \\
\leq &  \frac{c_2^4 \tau_n^2}{c_1^2} E[\lambda^{2}_{max}|\bm{Y}_n,\bm{X}_n] \cdot \| \bm{Y}_n-\bm{K}_n\bm{\beta}_{0n}\|_{2,n}^2 .
\end{split}\label{theorem1.1}
\end{equation}
\end{scriptsize}
\noindent
Next by the Schwarz inequality we have 
\begin{equation}
\begin{split}
& E_0[ \frac{c_2^4 \tau_n^2}{c_1^2} E[\lambda^{2}_{max}|\bm{Y}_n,\bm{X}_n] \cdot \| \bm{Y}_n-\bm{K}_n\bm{\beta}_{0n}\|_{2,n}^2]\\
\leq & \frac{c_2^4 \tau_n^2}{c_1^2}  E^{1/2}_0 (E^2[\lambda^{2}_{max}|\bm{Y}_n,\bm{X}_n]) E^{1/2}_0 [\| \bm{Y}_n-\bm{K}_n\bm{\beta}_{0n}\|_{2,n}^4] \\
\leq & \frac{c_2^4 \tau_n^2}{c_1^2}  E_0^{1/2}( E[\lambda^{4}_{max}|\bm{Y}_n,\bm{X}_n]) \cdot E_0 ^{1/2}(\chi^2_n)^2 \\
=& \frac{c_2^4 \tau_n^2}{c_1^2}  E_0^{1/2}( \lambda^{4}_{max}) \cdot \sqrt{ n(n+2)\sigma_0^4}\\
\leq & \sqrt{n} \frac{c_2^4 \tau_n^2}{c_1^2}  E_0^{1/2}( \lambda_1^{4}) \cdot \sqrt{ n(n+2)\sigma_0^4} ,
\end{split}\label{theorem1.2}
\end{equation}
By(\ref{theorem1.1}),(\ref{theorem1.2}), $E_0 \parallel I \parallel_{2,n}^2 \preceq q_n$.\\
\\
\\
Next we show $E_0\parallel II \parallel_{2,n}^2 \preceq q_n$. \\
\\
Since $\bm{I}_n \leq c_1^{-2}\bm{K}^2_n $,  $\bm{K}^2_n=\bm{K}_n\bm{I}_n\bm{K}_n \leq c_1^{-2}\bm{K}^4_n$,
\begin{equation}
\begin{split}
& E_0\parallel II \parallel_{2,n}^2 \\
 \leq & E_0E[\| \bm{K}_n(\bm{K}_n^2+\tau_n^{-2}\bm{\Lambda}^{-2}_n)^{-1}\tau_n^{-2}\bm{\Lambda}^{-2}_n\bm{\beta}_{0n} \|_{2,n}^2|\bm{Y}_n,\bm{X}_n]\\
\leq & {c^{-2}_1} E_0E[ \bm{\beta}^T_{0n}(\tau_n^{-2}\bm{\Lambda}^{-2}_n)(\bm{K}_n^2+\tau_n^{-2}\bm{\Lambda}^{-2}_n)^{-1}\bm{K}_n^4(\bm{K}_n^2+\tau_n^{-2}\bm{\Lambda}^{-2}_n)^{-1}(\tau_n^{-2}\bm{\Lambda}^{-2}_n)\bm{\beta}_{0n} |\bm{Y}_n,\bm{X}_n]  \\
= &  {c^{-2}_1} E_0E[ \bm{\beta}^T_{0n}(\bm{K}_n^{-2}+\tau_n^{2}\bm{\Lambda}^{2}_n)^{-2}\bm{\beta}_{0n} |\bm{Y}_n,\bm{X}_n] \\
= &  {c^{-2}_1} E_0E[ \bm{\beta}^T_{0n}(\bm{K}_n^{-2}+\tau_n^{2}\bm{\Lambda}^{2}_n)^{-1/2}(\bm{K}_n^{-2}+\tau_n^{2}\bm{\Lambda}^{2}_n)^{-1}(\bm{K}_n^{-2}+\tau_n^{2}\bm{\Lambda}^{2}_n)^{-1/2}\bm{\beta}_{0n} |\bm{Y}_n,\bm{X}_n] \\
\leq & c^{-2}_1 E_0E[ \bm{\beta}^T_{0n}(\bm{K}_n^{-2}+\tau_n^{2}\bm{\Lambda}^{2}_n)^{-1/2}\bm{K}_n^2 (\bm{K}_n^{-2}+\tau_n^{2}\bm{\Lambda}^{2}_n)^{-1/2}\bm{\beta}_{0n} |\bm{Y}_n,\bm{X}_n]\\
= &  (c^{2}_2 /c^{2}_1) E_0E[ \bm{\beta}^T_{0n}(\bm{K}_n^{-2}+\tau_n^{2}\bm{\Lambda}^{2}_n)^{-1} \bm{\beta}_{0n} |\bm{Y}_n,\bm{X}_n]\\
\leq &   (c^{2}_2 /c^{2}_1) E_0E[ \bm{\beta}^T_{0n}\bm{K}_n^{2} \bm{\beta}_{0n} |\bm{Y}_n,\bm{X}_n]
 \leq  ({c^4_2} /c^{2}_1 )\|\bm{\beta}_{0n}\|_{2,n}^2 \preceq q_n.
\end{split}
\end{equation}
The theorem follows.
\noindent
\\
\begin{remark} 
The assumption of finiteness of the second moment of $\lambda^2$ can be weakened. All we need is the finiteness of the $(1+\delta)$th moment of $\lambda^2$. To see this, one applies  Holder's inequality to get 
\begin{equation}
\begin{split}
& E_0[  E[\lambda^{2}_{max}|\bm{Y}_n,\bm{X}_n] \cdot \| \bm{Y}_n-\bm{K}_n\bm{\beta}_{0n}\|_{2,n}^2]\\
\leq &   E^{\frac{1}{1+\delta}}_0 (E^{1+\delta}[\lambda^{2}_{max}|\bm{Y}_n,\bm{X}_n]) E^{\frac{\delta}{1+\delta}}_0 [\| \bm{Y}_n-\bm{K}_n\bm{\beta}_{0n}\|_{2,n}^{2 \cdot \frac{1+\delta}{\delta}}] \\
\leq &  E^{\frac{1}{1+\delta}}_0 (E[\lambda^{2\cdot {(1+\delta)}}_{max}|\bm{Y}_n,\bm{X}_n]) E^{\frac{\delta}{1+\delta}}_0 [\| \bm{Y}_n-\bm{K}_n\bm{\beta}_{0n}\|_{2,n}^{2 \cdot \frac{1+\delta}{\delta}}]\\
= & E^{\frac{1}{1+\delta}} (\lambda^{2\cdot {(1+\delta)}}_{max}) E^{\frac{\delta}{1+\delta}}_0 [\| \bm{Y}_n-\bm{K}_n\bm{\beta}_{0n}\|_{2,n}^{2 \cdot \frac{1+\delta}{\delta}}].
\end{split} \label{rmk1}
\end{equation}
Since  $\| \bm{Y}_n-\bm{K}_n\bm{\beta}_{0n}\|_{2,n}^2 \sim \sigma_0^2 \chi_n^2$,
\begin{displaymath}
E_0 [\| \bm{Y}_n-\bm{K}_n\bm{\beta}_{0n}\|_{2,n}^{2 \cdot \frac{1+\delta}{\delta}}]=
(\sigma_0^2)^{\frac{1+\delta}{\delta}} (\chi_n^2)^{1+1/ \delta}=(2\sigma_0^2)^{\frac{1+\delta}{\delta}} \Gamma(n/2+1/\delta+1)/\Gamma(n/2).
\end{displaymath}
Using Stirling's formula, $\Gamma(n/2+1/\delta+1)/\Gamma(n/2) \leq C n^{1/\delta+1}$, so that the second term   in the right hand side of (\ref{rmk1}) is bounded above by a constant multiple of  $n$, also 
\begin{equation}
E (\lambda^{2\cdot {(1+\delta)}}_{max}) E^{\frac{1}{1+\delta}} (\lambda^{2\cdot {(1+\delta)}}_{max}) \leq n^{\frac{1}{1+\delta}} E^{\frac{1}{1+\delta}} (\lambda^{2\cdot {(1+\delta)}}_{1}) .\label{rmk1.2}
\end{equation}
By (\ref{rmk1}) and (\ref{rmk1.2}), 
\begin{equation}
 E_0[  E[\lambda^{2}_{max}|\bm{Y}_n,\bm{X}_n] \cdot \| \bm{Y}_n-\bm{K}_n\bm{\beta}_{0n}\|_{2,n}^2] \leq C n^{1+\frac{1}{1+\delta}} E^{\frac{1}{1+\delta}} (\lambda^{2\cdot {(1+\delta)}}_{1}).
\end{equation}
Then Theorem 1 holds with $\tau_n^2 \preceq n^{-1-\frac{1}{1+\delta}}q_n$.\end{remark}
\noindent
\\
\begin{remark}
 The assumption of $(1+\delta)$th moment of $\lambda^2$ holds for several distributions. Example include the common Gamma distribution, the inverse Gaussian distribution, Student's t-distribution with finite second moment, the inverse gamma distribution with shape parameter greater than $1+\delta$ and the beta prime priors $\pi(\lambda^2) \varpropto (\lambda^2)^{a-1}(1+\lambda^2)^{-a-b}$ with $b>1+\delta$.
 \end{remark}
\noindent
\begin{theorem} 
Let $K_n$ be a kernel with fixed parameter, and conditions (A1)-(A3) hold. Consider the  priors assigned to $\bm{\Lambda}_n$ and $\sigma^2$ as in Section 2.3.  Then if $\tau^2_n \preceq n^{-\frac{3}{2}}q_n$ and
\begin{displaymath}
\int \lambda^{4} \pi(\lambda^2)  d{\lambda^2} < \infty,
\end{displaymath}
  then
\begin{displaymath}
 E_0 \{tr [V(\bm{K}_n \bm{\beta}_{n}|\bm{Y}_n,\bm{X}_n)]\} \preceq q_n
\end{displaymath}
as $n \to \infty.$ \label{theorem2}
\end{theorem}
\textbf{Proof of Theorem \ref{theorem2}:} By (\ref{posterior 3}) we have 
$E(\sigma^2|\bm{\Lambda}_n,\bm{Y}_n,\bm{X}_n)=\frac{b+{\bm{Y}_n}^T{(\bm{I}_n-\bm{K}_n(\bm{K}_n^2+\tau_n^{-2}\bm{\Lambda}^{-2}_n)^{-1}\bm{K}_n)}{\bm{Y}_n}}{n+a-2},$
\begin{equation}
\begin{split}
& tr[V(\bm{K}_n\bm{\beta}_{n}|\bm{Y}_n,\bm{X}_n)] \\
= & tr E[V(\bm{K}_n\bm{\beta}_{n}|\sigma^2,\bm{\Lambda}_n,\bm{Y}_n,\bm{X}_n)|\bm{Y}_n,\bm{X}_n] \\
&+ tr V[E(\bm{K}_n\bm{\beta}_{n}|\sigma^2,\bm{\Lambda}_n,\bm{Y}_n,\bm{X}_n)|\bm{Y}_n,\bm{X}_n] \\
= & trE[\frac{b+{\bm{Y}_n}^T{(\bm{I}_n-\bm{K}_n(\bm{K}_n^2+\tau_n^{-2}\bm{\Lambda}^{-2}_n)^{-1}\bm{K}_n)}{\bm{Y}_n}}{n+a-2}\bm{K}_n(\bm{K}_n^2+\tau_n^{-2}\bm{\Lambda}^{-2}_n)^{-1}\bm{K}_n|\bm{Y}_n,\bm{X}_n]\\
& + tr V[\bm{K}_n(\bm{K}_n^2+\tau_n^{-2}\bm{\Lambda}^{-2}_n)^{-1}\bm{K}_n\bm{Y}_n|\bm{Y}_n,\bm{X}_n] \\
= & III+ IV, say.
\end{split} \label{theorem2.1}
\end{equation}

\begin{equation}
\begin{split}
III \le &  trE[\frac{b+{\bm{Y}_n}^T{\bm{Y}_n}}{n+a-2} (\tau_n^{2} \lambda^2_{max}\bm{K}^2_n)|\bm{Y}_n,\bm{X}_n]\\
\leq & nc^2_2\tau_n^{2} E[ \lambda^2_{max}|\bm{Y}_n,\bm{X}_n]\frac{b+{\bm{Y}_n}^T{\bm{Y}_n}}{n+a-2} \\
\leq & nc^2_2\tau_n^{2} E[ \lambda^2_{max}|\bm{Y}_n,\bm{X}_n]\frac{b+2{(\bm{Y}_n-\bm{K}_n\bm{\beta}_{0n})}^T{(\bm{Y}_n-\bm{K}_n\bm{\beta}_{0n})}+2\bm{\beta}_{0n}^T\bm{K}^2_n\bm{\beta}_{0n}}{n+a-2}.
\end{split} \label{theorem2.2}
\end{equation}

By the Schwarz inequality, 
\begin{equation}
\begin{split}
E_0 III \le & nc^2_2\tau_n^{2} E^{1/2}_0 \{E^2[\lambda^2_{max}|\bm{Y}_n,\bm{X}_n]\}\frac{b+2\bm{\beta}_{0n}^T\bm{K}^2_n\bm{\beta}_{0n}}{n+a-2} \\
+& 2nc^2_2\tau_n^{2} E^{1/2}_0 \{E^2[\lambda^2_{max}|\bm{Y}_n,\bm{X}_n]\} E^{1/2}_0 \{\frac{{(\bm{Y}_n-\bm{K}_n\bm{\beta}_{0n})}^T{(\bm{Y}_n-\bm{K}_n\bm{\beta}_{0n})}}{n+a-2}\}^2\\
\leq & nc^2_2\tau_n^{2} E^{1/2}_0 \{E\lambda^4_{max}\}\frac{b+2\bm{\beta}_{0n}^T\bm{K}^2_n\bm{\beta}_{0n}}{n+a-2} 
+  2 nc^2_2\tau_n^{2} E^{1/2}_0 \{E\lambda^4_{max}\} \frac{\sqrt{\sigma_0^4 n(n+2)}}{n+a-2}\\
\leq & n^{3/2}c^2_2\tau_n^{2} E^{1/2}_0 \{E\lambda^4_{1}\}\frac{b+2q_n c_2^2}{n+a-2} 
+  2 n^{3/2} c^2_2\tau_n^{2} E^{1/2}_0 \{E\lambda^4_{1}\} \frac{\sqrt{\sigma_0^4 n(n+2)}}{n+a-2}\\
\preceq & q_n .
\end{split}\label{theorem2.3}
\end{equation}

\begin{equation}
\begin{split}
IV & = tr V[\bm{K}_n(\bm{K}_n^2+\tau_n^{-2}\bm{\Lambda}^{-2}_n)^{-1}\bm{K}_n(\bm{Y}_n-\bm{K}_n\bm{\beta}_{0n}+\bm{K}_n\bm{\beta}_{0n})|\bm{Y}_n,\bm{X}_n]\\
& \le 2 tr V[\bm{K}_n(\bm{K}_n^2+\tau_n^{-2}\bm{\Lambda}^{-2}_n)^{-1}\bm{K}_n(\bm{Y}_n-\bm{K}_n\bm{\beta}_{0n})|\bm{Y}_n,\bm{X}_n] \\
&+ 2 tr V[\bm{K}_n(\bm{K}_n^2+\tau_n^{-2}\bm{\Lambda}^{-2}_n)^{-1}\bm{K}_n^2\bm{\beta}_{0n}|\bm{Y}_n,\bm{X}_n]. \label{theorem2.4}
\end{split}
\end{equation}
\noindent
The 1st term in the RHS of (\ref{theorem2.4}) 
\begin{displaymath}
\begin{split}
\le  2 E[\| \bm{K}_n(\bm{K}_n^2+\tau_n^{-2}\bm{\Lambda}^{-2}_n)^{-1}\bm{K}_n(\bm{Y}_n-\bm{K}_n\bm{\beta}_{0n})\|^2_{2,n}|\bm{Y}_n,\bm{X}_n] .
\end{split}
\end{displaymath}
Then by (\ref{theorem1.1}), (\ref{theorem1.2}),we have 
\begin{equation}
E_0 \{tr V[\bm{K}_n(\bm{K}_n^2+\tau_n^{-2}\bm{\Lambda}^{-2}_n)^{-1}\bm{K}_n(\bm{Y}_n-\bm{K}_n\bm{\beta}_{0n})|\bm{Y}_n,\bm{X}_n]\}  \preceq q_n .\label{theorem2.5}
\end{equation}
\noindent
The 2nd term in the RHS of (\ref{theorem2.4}) 
\begin{displaymath}
\begin{split}
\leq & 2 E[ \| \bm{K}_n(\bm{K}_n^2+\tau_n^{-2}\bm{\Lambda}^{-2}_n)^{-1}\bm{K}_n^2\bm{\beta}_{0n}\|^2_{2,n}|\bm{Y}_n,\bm{X}_n] \\
\le & 2 c_2^2 \bm{\beta}_{0n}^T  \bm{K}_n^2   (\bm{K}_n^2+\tau_n^{-2}\bm{\Lambda}^{-2}_n)^{-2}  \bm{K}_n^2   \bm{\beta}_{0n}\\ 
= & 2 c_2^2 \bm{\beta}_{0n}^T  \bm{K}_n^2   (\bm{K}_n^2+\tau_n^{-2}\bm{\Lambda}^{-2}_n)^{-1/2} (\bm{K}_n^2+\tau_n^{-2}\bm{\Lambda}^{-2}_n)^{-1} (\bm{K}_n^2+\tau_n^{-2}\bm{\Lambda}^{-2}_n)^{-1/2} \bm{K}_n^2   \bm{\beta}_{0n}   \\
\leq &  2 (c_2^2/c_1^2) \bm{\beta}_{0n}^T  \bm{K}_n^2  (\bm{K}_n^2+\tau_n^{-2}\bm{\Lambda}^{-2}_n)^{-1}  \bm{K}_n^2   \bm{\beta}_{0n}     \\
\leq &  2 (c_2^2/c_1^2) \bm{\beta}_{0n}^T  \bm{K}_n^2  (\bm{K}_n^2)^{-1}  \bm{K}_n^2   \bm{\beta}_{0n}     \\
\leq & 2 \frac{c_2^4}{c_1^2} \| \bm{\beta}_{0n}\|^2_{2,n}\preceq q_n. 
\end{split}
\end{displaymath}
Hence,
\begin{equation}
E_0 \{tr V[\bm{K}_n(\bm{K}_n^2+\tau_n^{-2}\bm{\Lambda}^{-2}_n)^{-1}\bm{K}_n^2\bm{\beta}_{0n}|\bm{Y}_n,\bm{X}_n]\} \preceq q_n .\label{theorem2.6}
\end{equation}
The theorem follows from (15)-(20). \\
\\
\\
\\
By Theorems 1 and 2, we immediately get 
\begin{corollary} 
If $K_n$ is a kernel with fixed parameter, and conditions (A1)-(A3) hold,  with the prior assigned to $\bm{\Lambda}_n$ and $\sigma^2$ in model 1, if $\tau^2_n \preceq n^{-\frac{3}{2}}q_n$ and
\begin{displaymath}
\int \lambda^{4} \pi(\lambda^2)  d{\lambda^2} < \infty,
\end{displaymath}
  then
\begin{displaymath}
 E_0 P(\| \bm{K}_n \bm{\beta}_{n}-\bm{K}_n \bm{\beta}_{0n} \|^2_{2,n} \ge M_nq_n |\bm{Y}_n,\bm{X}_n) \to 0
\end{displaymath}
as $n \to \infty$, where $M_n\rightarrow\infty$ as $\rightarrow\infty$.  
\end{corollary}
In particular, one may take $M_n=\log (n/q_n)$ to get the asymptotic minimax contraction bound.

\section{Hierarchical Bayesian Model with Polynomial  Kernel}
For a polynomial kernel, we can not apply Theorem \ref{theorem1} directly, because the regularity condition (A1) does not generally  hold. 
For example, for polynomial kernel with fixed parameter $\theta >0$, if  the design matrix $\bm{X}_n$ satisfies the orthogonality condition, namely,
 $\bm{X}_n\bm{X}_n^T=p\bm{I}_n$, then $p^{-\theta}K_n \to I_n$ as $n\rightarrow\infty$ so that  $\frac{1}{2} p^{\theta}\bm{I}_n\le \bm{K}_n \le 2 p^{\theta}\bm{I}_n$ for sufficiently large $n$, which does not satisfy condition (A1). In this section, we consider the case when the design matrix $\bm{X}_n$ satisfies  $t_1(n) \bm{I}_n \le \bm{K}_n \le t_2(n)\bm{I}_n$, $t_2(n)/{t_1(n)}=O(1)$, where $t_1(n)$ and $t_2(n)$ are functions  depending solely on $n$. \\
 \\
\\
 We have the following posterior contraction result for  polynomial kernels.
\begin{theorem} 
If $K_n$ is a kernel with fixed parameter, and conditions (A2),(A3) hold,  with the prior assigned to $\bm{\Lambda}_n$ and $\sigma^2$ in model 1, if $t_1(n) \bm{I}_n \le \bm{K}_n \le t_2(n)\bm{I}_n$, $t_2(n)/{t_1(n)}=O(1)$, $t^2_2(n) \prec  n q_n^{-1}$ ,  $\tau_n^2=q_n n^{-3/2}$ and
\begin{displaymath}
\int \lambda^{4} \pi(\lambda^2)  d{\lambda^2} < \infty,
\end{displaymath}
  then
\begin{displaymath}
 E_0\parallel E(\bm{K}_n \bm{\beta}_{n}|\bm{Y}_n,\bm{X}_n)-\bm{K}_n \bm{\beta}_{0n}\parallel_{2,n}^2 \preceq q_n t_2^2(n) 
\end{displaymath}
as $n \to \infty.$   \label{theorem3}
\end{theorem}
\textbf{Proof of Theorem \ref{theorem3}:} This proof is almost the same as of Theorem \ref{theorem1}. It suffices to show that $E_0 \parallel I \parallel_{2,n}^2 \preceq q_n t_2^2(n) $  and $E_0\parallel II \parallel_{2,n}^2 \preceq q_n t_2^2(n) $  as $n \to \infty$.\\
\\
Substituting $c_1$ and $c_2$ in Theorem \ref{theorem1} by $t_1(n)$ and $t_2(n)$, we get
\begin{equation}
E_0\parallel I \parallel_{2,n}^2 \leq \sqrt{n} \frac{t^4_2(n) \tau_n^2}{t^2_1(n)}  E_0^{1/2}( \lambda_1^{4}) \cdot \sqrt{ n(n+2)\sigma_0^4} \preceq q_n t_2^2(n) 
\end{equation}
and
\begin{equation}
E_0\parallel II \parallel_{2,n}^2 \leq  ({t^4_2(n)} /t^{2}_1(n) )\|\bm{\beta}_{0n}\|_{2,n}^2 \preceq  q_n t_2^2(n) .
\end{equation}

\noindent
\begin{theorem} 
Let $K_n$ be a kernel with fixed parameter, and conditions (A2) and (A3) hold.  Then with the same priors assigned to $\bm{\Lambda}_n$ and $\sigma^2$ in Section 2.2, if $t_1(n) \bm{I}_n \le \bm{K}_n \le t_2(n)\bm{I}_n$, $t_2(n)/{t_1(n)}=O(1)$, $t^2_2(n) \prec  n q_n^{-1}$ ,  $\tau_n^2=q_n n^{-3/2}$  an
$\int \lambda^{4} \pi(\lambda^2)  d{\lambda^2} < \infty$, then
\begin{displaymath}
 E_0 \{tr [V(\bm{K}_n \bm{\beta}_{n}|\bm{Y}_n,\bm{X}_n)]\} \preceq q_n t_2^2(n) 
\end{displaymath}
as $n \to \infty.$  \label{theorem4}
\end{theorem}
\textbf{Proof of Theorem \ref{theorem4}:} This proof is almost the same as of theorem \ref{theorem2}. Substitute $c_1$, $c_2$ in theorem \ref{theorem2} by $t_1(n)$, $t_2(n)$. 
\\
\\
\\
Combining  Theorems \ref{theorem3} and \ref{theorem4} we get
\begin{corollary} 
Let $K_n$ be a kernel with fixed parameter, and conditions (A2) and (A3) hold.  Consider the same priors assigned to $\bm{\Lambda}_n$ and $\sigma^2$ as in Section 2.2. If $t_1(n) \bm{I}_n \le \bm{K}_n \le t_2(n)\bm{I}_n$, $t_2(n)/{t_1(n)}=O(1)$, $t^2_2(n) \prec  n q_n^{-1}$ ,  $\tau_n^2=q_n n^{-3/2}$  and
\begin{displaymath}
\int \lambda^{4} \pi(\lambda^2)  d{\lambda^2} < \infty,
\end{displaymath}
  then
\begin{displaymath}
 E_0 P(\| \bm{K}_n \bm{\beta}_{n}-\bm{K}_n \bm{\beta}_{0n} \|^2_{2,n} \ge q_n t_2^2(n)  |\bm{Y}_n,\bm{X}_n) \to 0
\end{displaymath}
as $n \to \infty.$ \label{corollary2}
\end{corollary}
\noindent
\\
\begin{remark}  
For polynomial kernels with fixed parameters, if the design matrix is approximately orthogonal , then we still have  $t_1(n) \bm{I}_n \le \bm{K}_n \le t_2(n)\bm{I}_n$, $t_2(n)/{t_1(n)}=O(1)$.  Theorems \ref{theorem3} , \ref{theorem4} and Corollary \ref{corollary2} hold if $p^{\theta} \preceq \sqrt{nq_n^{-1}}$.  Actually we have the following lemma which describes the behavior of the kernel $K_n$ when the design matrix is approximately orthogonal. 
\begin{lemma}
Let $K_n$ be a polynomial kernel with parameter $\theta \in [a_L, a_U]$, $a_L >1/2$, and  the design matrix $\bm{X}_n$ satisfies  
\begin{displaymath}
|\frac{\bm{x}_i\cdot\bm{x}_i+1}{p}-1|\le\frac{1}{h(n)},|\frac{\bm{x}_i\cdot\bm{x}_j+1}{p}|\le \frac{1}{k(n)},1\le i \neq j \le n,
\end{displaymath}
$h(n)=2 a_U \cdot n$, $k(n)=n^4$, for sufficiently large $n$. 
Then there exists $N >0$ such that when $n>N$, $(1-\frac{1}{n})\bm{I}_n \le p^{-\theta}\bm{K}_n \le (1+\frac{1}{n})\bm{I}_n$  for all $\theta \in[a_L,a_U]$.\label{lemma3.1}
\end{lemma}
\textbf{Proof:} It suffices to show that  for every $\bm{c} \neq \bm{0}$, $\bm{c}^T(p^{-\theta}\bm{K}_n)\bm{c} \le (1+\frac{1}{n})\bm{c}^T\bm{c}$ for large $n$.
But 
\begin{displaymath}
\begin{split}
&\bm{c}^T(p^{-\theta}\bm{K}_n)\bm{c} \\
\le & \Sigma_{i=1}^{n}{c_i}^2(1+\frac{1}{h(n)})^{\theta}+\Sigma_{1\le i \neq j \le n}|c_i||c_j|/k^{\theta}(n) \\
= & (1+\frac{1}{h(n)})^{\theta}\Sigma_{i=1}^{n}{c_i}^2+[(\Sigma_{i=1}^{2}{|c_i|})^2-\Sigma_{i=1}^{2}{c_i}^2]/(2k^{\theta}(n))\\
\le & [(1+\frac{1}{h(n)})^{\theta}-\frac{1}{2k^{\theta}(n)}]\Sigma_{i=1}^{n}{c_i}^2
+\frac{n\Sigma_{i=1}^{2}{c_i}^2}{2k^{\theta}(n)}\\
\le &[(1+\frac{1}{h(n)})^{\theta}-\frac{1}{2n^{4\theta}}+\frac{1}{2n^{4\theta -1}}]
\Sigma_{i=1}^{n}{c_i}^2\\
\le &  (1+\frac{1}{n})\bm{c}^T\bm{c}, \textrm{    (since  $\theta > \frac{1}{2}$)}.
\end{split}
\end{displaymath}
Similarly, we have $\bm{c}^T(p^{-\theta}\bm{K}_n)\bm{c} \ge (1-\frac{1}{n})\bm{c}^T\bm{c}$.\\
\end{remark}
\noindent
Although we have to assume $p<n$ to get posterior contraction results  for polynomial kernels, we can still get posterior consistency for the $p>n$ case for polynomial kernels. 
\begin{theorem} 
Let $K_n$ be a kernel with fixed parameter, and conditions (A2) and(A3) hold.   Consider the same priors assigned to $\bm{\Lambda}_n$ and $\sigma^2$ as in Section 2.2,  If $t_1(n) \bm{I}_n \le \bm{K}_n \le t_2(n)\bm{I}_n$, $t_2(n)/{t_1(n)}=O(1)$, if $t_1(n) \succ \sqrt{q_n n^{3/2}}$,  $\tau^2_n = \sqrt{q_nt_1^{-2}(n)n^{-\frac{3}{2}}} $ ,
\begin{displaymath}
\int \lambda^{4} \pi(\lambda^2)  d{\lambda^2} < \infty  , \qquad \int \lambda^{-2} \pi(\lambda^2)  d{\lambda^2} < \infty,
\end{displaymath}
  then
\begin{displaymath}
 E_0 P(\|  \bm{\beta}_{n}- \bm{\beta}_{0n} \|^2_{2,n} \ge  \sqrt{q_nt_1^{-2}(n)n^{\frac{3}{2}}} |\bm{Y}_n,\bm{X}_n) \to 0
\end{displaymath}
as $n \to \infty.$ \label{theorem5}
\end{theorem}
\begin{remark} 
For polynomial kernels, condition $t_1(n) \succ \sqrt{q_n n^{3/2} }$ means $p \succ  (q_n n^{3/2} )^{\frac{1}{2\theta}}$.
\end{remark}
\noindent
\\
\textbf{Proof of Theorem \ref{theorem5}:} We prove this theorem by combining the next two lemmas.\\
\begin{lemma} 
Let $K_n$ is a kernel with a fixed parameter, and assume conditions (A2) and (A3) to hold,  Then with the prior assigned to $\bm{\Lambda}_n$ and $\sigma^2$ in the  model given in Section 2.2, if $t_1(n) \bm{I}_n \le \bm{K}_n \le t_2(n)\bm{I}_n$, $t_2(n)/{t_1(n)}=O(1)$,  $t_1(n) \succ \sqrt{q_n n^{3/2}}$,  $\tau^2_n = \sqrt{q_nt_1^{-2}(n)n^{-\frac{3}{2}}} $,
\begin{displaymath}
\int \lambda^{4} \pi(\lambda^2)  d{\lambda^2} < \infty  , \qquad \int \lambda^{-2} \pi(\lambda^2)  d{\lambda^2} < \infty ,
\end{displaymath}
  then
\begin{displaymath}
 E_0\parallel E(\bm{\beta}_{n}|\bm{Y}_n,\bm{X}_n)-\bm{\beta}_{0n}\parallel_{2,n}^2 \preceq \sqrt{q_nt_1^{-2}(n)n^{\frac{3}{2}}}
\end{displaymath}
as $n \to \infty.$  \label{lemma3}
\end{lemma}
\textbf{Proof of Lemma \ref{lemma3}:} Since we have 
\begin{equation}
\begin{split}
E(\bm{\beta}_n|\bm{Y}_n,\bm{X}_n)-\bm{\beta}_{0n}
=&E(E(\bm{\beta}_n|\sigma^2,\bm{\Lambda}^2_n,\bm{Y}_n,\bm{X}_n)|\bm{Y}_n,\bm{X}_n)-\bm{\beta}_{0n}\\
=& E[(\bm{K}_n^2+\tau_n^{-2}\bm{\Lambda}^{-2}_n)^{-1}\bm{K}_n(\bm{Y}_n-\bm{K}_n\bm{\beta}_{0n})|\bm{Y}_n,\bm{X}_n] \\
& -E[(\bm{K}_n^2+\tau_n^{-2}\bm{\Lambda}^{-2}_n)^{-1}\tau_n^{-2}\bm{\Lambda}^{-2}_n\bm{\beta}_{0n}|\bm{Y}_n,\bm{X}_n]\\
= & V_1-V_2(say).
\end{split} 
\end{equation}
\noindent
\\
It suffices to show that $E_0 \parallel V_1 \parallel_{2,n}^2 \preceq  \sqrt{q_nt_1^{-2}(n)n^{\frac{3}{2}}} $  and $E_0\parallel V_2 \parallel_{2,n}^2 \preceq \sqrt{q_nt_1^{-2}(n)n^{\frac{3}{2}}} $  as $n \to \infty$.\\
\\
First, consider $E_0 \parallel V_1 \parallel_{2,n}^2 \preceq  \sqrt{q_nt_1^{-2}(n)n^{\frac{3}{2}}} $.  Similar to (\ref{theorem1.1}),(\ref{theorem1.2}) in the proof of theorem \ref{theorem1}, we get 
\begin{equation}
E_0 \parallel V_1 \parallel_{2,n}^2 \leq \sqrt{n} \frac{t_2^2(n) \tau_n^2}{t_1^2(n)}  E_0^{1/2}( \lambda_1^{4}) \cdot \sqrt{ n(n+2)\sigma_0^4} \preceq \sqrt{q_nt_1^{-2}(n)n^{\frac{3}{2}}}.
\end{equation}
Next we show $E_0\parallel VI \parallel_{2,n}^2 \preceq \sqrt{q_nt_1^{-2}(n)n^{\frac{3}{2}}}$. \\
\\
\begin{equation}
\begin{split}
& E_0\parallel V_2 \parallel_{2,n}^2 \\
 \leq & E_0E[\| (\bm{K}_n^2+\tau_n^{-2}\bm{\Lambda}^{-2}_n)^{-1}\tau_n^{-2}\bm{\Lambda}^{-2}_n\bm{\beta}_{0n} \|_{2,n}^2|\bm{Y}_n,\bm{X}_n]\\
= &  E_0E[ \bm{\beta}^T_{0n}(\tau_n^{-2}\bm{\Lambda}^{-2}_n)(\bm{K}_n^2+\tau_n^{-2}\bm{\Lambda}^{-2}_n)^{-1/2}(\bm{K}_n^2+\tau_n^{-2}\bm{\Lambda}^{-2}_n)^{-1}(\bm{K}_n^2+\tau_n^{-2}\bm{\Lambda}^{-2}_n)^{-1/2}(\tau_n^{-2}\bm{\Lambda}^{-2}_n)\bm{\beta}_{0n} |\bm{Y}_n,\bm{X}_n]  \\
\leq & 1/t_1^2(n) E_0E[ \bm{\beta}^T_{0n}(\tau_n^{-2}\bm{\Lambda}^{-2}_n)(\bm{K}_n^2+\tau_n^{-2}\bm{\Lambda}^{-2}_n)^{-1}(\tau_n^{-2}\bm{\Lambda}^{-2}_n)\bm{\beta}_{0n} |\bm{Y}_n,\bm{X}_n] \\
\leq & 1/t_1^2(n) E_0E[ \bm{\beta}^T_{0n}(\tau_n^{-2}\bm{\Lambda}^{-2}_n)\bm{\beta}_{0n} |\bm{Y}_n,\bm{X}_n] \\
\leq &   \frac{M^2 q_n}{t_1^2(n)\tau_n^{2}} E_0E[ \lambda_1^{-2} |\bm{Y}_n,\bm{X}_n]
= \frac{M^2 q_n}{t_1^2(n)\tau_n^{2}} E(\lambda_1^{-2})  \preceq  \sqrt{q_nt_1^{-2}(n)n^{\frac{3}{2}}}.
\end{split}
\end{equation}

\begin{lemma} 
If $K_n$ is a kernel with fixed parameter, and conditions (A2),(A3) hold,  with the prior assigned to $\bm{\Lambda}_n$ and $\sigma^2$ in model 1, if $t_1(n) \bm{I}_n \le \bm{K}_n \le t_2(n)\bm{I}_n$, $t_2(n)/{t_1(n)}=O(1)$, if $t_1(n) \succ \sqrt{q_n n^{3/2}}$,  $\tau^2_n = \sqrt{q_nt_1^{-2}(n)n^{-\frac{3}{2}}} $ and
\begin{displaymath}
\int \lambda^{4} \pi(\lambda^2)  d{\lambda^2} < \infty  , \qquad \int \lambda^{-2} \pi(\lambda^2)  d{\lambda^2} < \infty
\end{displaymath}
  then
\begin{displaymath}
 E_0 \{tr [V( \bm{\beta}_{n}|\bm{Y}_n,\bm{X}_n)]\} \preceq \sqrt{q_nt_1^{-2}(n)n^{\frac{3}{2}}}
\end{displaymath}
as $n \to \infty.$  \label{lemma4}
\end{lemma}
\textbf{Proof of Lemma \ref{lemma4}:} By the same method in theorem \ref{theorem2}, we can prove  this lemma.  There is only one step is slightly different from theorem \ref{theorem2}, equation (\ref{theorem2.6}).
\begin{equation}
\begin{split}
& E_0 \{tr V[(\bm{K}_n^2+\tau_n^{-2}\bm{\Lambda}^{-2}_n)^{-1}\bm{K}_n^2\bm{\beta}_{0n}|\bm{Y}_n,\bm{X}_n]\} \\
= & E_0 \{trV[(\bm{K}_n^2+\bm{\Lambda}^{-2}_n)^{-1}\bm{\Lambda}^{-2}_n\bm{\beta}_{0n}|\bm{Y}_n,\bm{X}_n]\}\\
\le & E_0  \{tr E[(\bm{K}_n^2+\bm{\Lambda}^{-2}_n)^{-1}\bm{\Lambda}^{-2}_n\bm{\beta}_{0n}\bm{\beta}_{0n}^{-T}\bm{\Lambda}^{-2}_n(\bm{K}_n^2+\bm{\Lambda}^{-2}_n)^{-1}|\bm{Y}_n,\bm{X}_n]\} \\
= & E_0 \{\bm{\beta}_{0n}^{-T}E[\bm{\Lambda}^{-2}_n(\bm{K}_n^2+\bm{\Lambda}^{-2}_n)^{-2}\bm{\Lambda}^{-2}_n|\bm{Y}_n,\bm{X}_n]\bm{\beta}_{0n}\} \\
\le &E_0 \{\bm{\beta}_{0n}^{-T}E[\bm{\Lambda}^{-2}_n  (\bm{K}_n^2+\bm{\Lambda}^{-2}_n)^{-1/2}(t_1^2(n)\bm{I}_n)^{-1}(\bm{K}_n^2+\bm{\Lambda}^{-2}_n)^{-1/2}\bm{\Lambda}^{-2}_n|\bm{Y}_n,\bm{X}_n]\bm{\beta}_{0n}\}\\
\le & t_1^{-2}(n) E_0 \{\bm{\beta}_{0n}^{-T}E[\bm{\Lambda}^{-2}_n|\bm{Y}_n,\bm{X}_n]\bm{\beta}_{0n}\}\\
\leq & t_1^{-2}(n) M^2 q_n E_0 \{E[{\lambda}^{-2}_1|\bm{Y}_n,\bm{X}_n]\\
= & t_1^{-2}(n) M^2 q_n E({\lambda}^{-2}_1) \preceq \sqrt{q_nt_1^{-2}(n)n^{\frac{3}{2}}}.
\end{split}
\end{equation}

\section{Discussion}
\citet{Tipping:2001} pointed out that RVM is a Gaussian  process model.  \citet{Vaart:2008} and \citet{Ghosal:2017},  obtained several posterior concentration results for Gaussian process models. They considered estimating a regression function $f$ based on observations $\bm{Y}_{1n}, \cdots, \bm{Y}_{nn}$
in a normal regression model with fixed covariates $\bm{Y}_{in}=f(\bm{x}_i)+\epsilon_i$, where $\epsilon_i  \stackrel{i.i.d}{\sim} N(0, \sigma^2_0)$ and the covariates $\bm{x}_1, \cdots, \bm{x}_n$ are fixed elements from a set $\mathcal{X}$. \\
\\
A prior on $f$ is induced by setting $f(\bm{x})=W_{\bm{x}}$, for a Gaussian process $(W_{\bm{x}}:x \in \mathcal{X})$. Any Gaussian element in a separable Banach space can be expanded as an infinite series $\sum_i Z_i h_i$ for i.i.d standard normal variables $Z
_i $ and elements $h_i$ from its RKHS. \citet{Vaart:2008} truncated this infinite series at a sufficient high level  to get a new Gaussian process prior.  If this  truncated   series converges  to the 
infinite series quickly, then by Theorem 2.2 in \citet{Vaart:2008},  the same posterior rate of contraction is attained. 
Since finite sums may be easier to handle, it is interesting to investigate special expansions and the number of terms that need to be retained in order to obtain the same contraction rate. 
\citet{Vaart:2008} illustrated  this  by an example of  the truncated wavelet expansion of functions in 
$\mathbb{L}_2([0,1]^d)$.  \citet{Ghosal:2007}  considered the truncated B-spline expansion in their Theorem 12.   These truncated series are quite similar to our model if  set $p=d$ fixed , $\bm{	K}_n=\bm{I}_n$ and the prior $ \bm{\beta}_n \sim N(\mathbf{0},\bm{I}_n)$. However in their case,  the number of terms  in the random series is $O(n^{\alpha})$, $\alpha<1$,  while ours is $n$.  Adding global shrinkage parameter $\tau_n $ to accommodate sparsity seems reasonable. Also, the Gaussian process prior related to RVM is data dependent, which is likely to add flexibility  to prediction.\\
\\
In our model, we assume the true model  can be write explicitly as $f_0(\bm{x})=\Sigma_{j=1}^{n}\beta_jK(\bm{x} ,\bm{x}_j)$, which is dependent on the $x_i$.  A natural question is how a posterior contraction result holds if $\Sigma_{j=1}^{n}\beta_jK(\bm{x} ,\bm{x}_j)$ is just a good approximation of $f_0(\bm{x})$ like the condition (7.10) in \citet{Ghosal:2007} .\\
\\
Also, the regularity condition (A2) is a  strong assumption. One needs to investigate a possible  posterior contraction result without this condition.  \\

\section*{Acknowledgments}
We would like to thank  the referee for his many valuable comments
which improved the paper considerably.

\vskip 0.2in
\bibliography{sample}

\end{document}